\pdfoutput=1
\documentclass[11pt]{article}

\usepackage[margin=1in]{geometry}
\usepackage{times}
\usepackage{microtype}

\usepackage{amsmath,amssymb,amsthm,mathtools}
\usepackage{bm}

\usepackage{graphicx}
\usepackage{booktabs}
\usepackage{tikz}

\usepackage[hidelinks]{hyperref}
\usepackage[nameinlink,capitalize]{cleveref}
\usepackage[numbers]{natbib}

\usepackage{algorithm}
\usepackage{algpseudocode}

\usepackage{enumitem}
\setlist{nosep}

\newtheorem{proposition}{Proposition}

\usepackage[numbers]{natbib}

\title{Neural Field Turing Machine: A Differentiable Spatial Computer}
\author{%
  \textbf{Akash Malhotra, Nacéra Seghouani}\\
  \texttt{akash.malhotra@lisn.fr, nacera.seghouani@lisn.fr}
}
\date{\today}

\begin{document}
\maketitle

\begin{abstract}
We introduce the Neural Field Turing Machine (NFTM), a differentiable architecture that unifies symbolic computation, physical simulation, and perceptual inference within continuous spatial fields. NFTM combines a neural controller, continuous memory field, and movable read/write heads that perform local updates. At each timestep, the controller reads local patches, computes updates via learned rules, and writes them back while updating head positions. This design achieves linear O(N) scaling through fixed-radius neighborhoods while maintaining Turing completeness under bounded error. We demonstrate three example instantiations of NFTM: cellular automata simulation (Rule 110), physics-informed PDE solvers (2D heat equation), and iterative image refinement (CIFAR-10 inpainting). These instantiations learn local update rules that compose into global dynamics, exhibit stable long-horizon rollouts, and generalize beyond training horizons. NFTM provides a unified computational substrate bridging discrete algorithms and continuous field dynamics within a single differentiable framework.
\end{abstract}

\section{Introduction}
\label{sec:introduction}
Many of the hardest problems in computer vision, robotics, and physics involve reasoning over \emph{spatially continuous fields}. 
From fluid dynamics and weather forecasting to scene understanding and robot control, the world presents itself not as a sequence of tokens but as multi-dimensional spatial fields that evolve over time. 
Despite their successes, current machine learning architectures are poorly matched to this setting. 
Transformers excel at sequence modeling but treat space as discrete tokens and suffer from quadratic attention cost. 
Diffusion models perform powerful iterative refinements but rely on stochastic noise injection rather than controllable update rules. 
Physics-informed approaches such as PINNs~\cite{raissi2019physics} and neural operators~\cite{li2020fourier,lu2019deeponet} approximate PDE dynamics but do not generalize beyond physics domains. 
Neural Cellular Automata (NCA)~\cite{mordvintsev2020growing} show the potential of local differentiable updates, yet lack an explicit controller or the flexibility to move beyond fixed-grid evolution. 
In short, existing models either sacrifice spatial continuity, locality, or algorithmic generality.  

We propose the \textit{Neural Field Turing Machine (NFTM)}, a neural architecture that embeds computation directly into a continuous field with differentiable read/write access. 
NFTM combines three ingredients: a neural controller, a spatial memory field, and movable read/write heads. 
At each timestep, the controller reads local patches of the field, computes updates via learned transition rules, and writes them back, simultaneously updating head positions. 
This formulation unifies symbolic and continuous computation: unlike Transformers, NFTM scales linearly with the number of discretized field sites $N$ through fixed-radius neighborhoods; unlike NCAs, NFTM has explicit controller dynamics and Turing completeness under bounded error; unlike diffusion models, NFTM produces deterministic rollouts at inference given fixed parameters and initial conditions; and unlike PINNs, NFTM is a general-purpose framework that extends naturally beyond PDEs.

The spatial nature of NFTM is central. 
Because it operates directly on continuous fields, NFTM is well-suited to problems where locality, geometry, and iterative simulation matter. 
Research in cognitive science suggests that aspects of human cognition, such as intuitive physics~\cite{battaglia2013simulation} and mental imagery~\cite{shepard1971mental}, involve running internal simulations to predict outcomes. 
NFTM echoes this paradigm by providing a differentiable substrate for simulation-based reasoning, bridging symbolic cellular automata, physical PDE solvers, and perceptual iterative inference within a single model.

Our work makes the following contributions:
\begin{itemize}
    \item We formally define the Neural Field Turing Machine (NFTM), a neural controller operating over continuous fields with explicit read/write heads, and prove its Turing completeness under bounded error.
    \item We demonstrate NFTM across three domains: symbolic (cellular automata), physical (heat equation with global and variable diffusion coefficients), and perceptual (image inpainting).
    \item We show that NFTM subsumes Neural Cellular Automata (NCA) as a special case, and provide an alternative to PINNs and iterative perceptual refinement models, positioning NFTM as a unifying framework for discrete and continuous computation.

\end{itemize}

Code and experiments are available at \url{https://github.com/akashjorss/NFTM}.

NFTM thus provides a blueprint for learned, differentiable, continuous-memory computers, a step toward architectures that unify algorithmic reasoning with the spatial and iterative structure of the physical world.

\section{Related Work}
\label{sec:related-work}
NFTM builds on a broad body of work spanning implicit neural representations, neural cellular automata, memory-augmented models, physics-informed networks, neural operators, graph neural networks, and diffusion models. We highlight the most relevant connections below.  

\paragraph{Implicit Neural Representations and Neural Fields.}
Coordinate-based models such as SIREN~\cite{sitzmann2020siren} and Fourier feature networks~\cite{tancik2020fourier} have demonstrated that neural networks can represent signals in continuous domains. Neural Radiance Fields (NeRF)~\cite{mildenhall2020nerf} and their extensions such as D-NeRF~\cite{pumarola2021dnerf}, Nerfies~\cite{park2020nerfies}, and Control-NeRF~\cite{lazova2022control} further show how neural fields can encode dynamic 3D structure. While these models learn static mappings from coordinates to values, NFTM instead operates through \emph{iterative updates} with a neural controller, enabling both discrete and continuous computation.

\paragraph{Neural Cellular Automata.}
Neural Cellular Automata (NCA)~\cite{mordvintsev2020growing, bigi2025universal, piguet2025cells2pixels} have shown that local differentiable update rules can generate complex emergent patterns, from self-organizing textures to universal computation. However, NCAs operate only over discrete grids with fixed neighborhoods. NFTMs subsume NCA dynamics as a special case ($\Delta h_t = 0$, fixed $A_t$), but add explicit read/write heads and controller logic, granting both locality and Turing-complete expressivity.

\paragraph{Neural Turing Machines and Memory-Augmented Models.}
Neural Turing Machines~\cite{graves2014neural} and Differentiable Neural Computers~\cite{graves2016hybrid} augment RNNs with differentiable memory, enabling algorithmic tasks such as copying and sorting. However, their memory is organized as a discrete matrix and indexed by soft attention, making them poorly suited to spatially continuous domains. NFTMs generalize this idea by embedding the memory in a \emph{continuous field}, providing differentiable read/write while retaining access to geometry and physics.

\paragraph{Physics-Informed Neural Networks and Neural Operators.}
Physics-Informed Neural Networks (PINNs)~\cite{raissi2019physics} solve PDEs by embedding the residuals of governing equations directly into the loss. Neural ODEs~\cite{chen2018neural}, Hamiltonian Neural Networks~\cite{greydanus2019hamiltonian}, and latent SDEs~\cite{li2020latent} extend this idea to learning continuous-time dynamics under expressive priors. Neural operators such as DeepONet~\cite{lu2019deeponet} and Fourier Neural Operators (FNO)~\cite{li2020fourier} learn mappings between infinite-dimensional function spaces. NFTMs differ by implementing forward Euler-style updates on a spatial field, keeping computation local and explicit like classical numerical solvers, but fully differentiable and trainable end-to-end.

\paragraph{Graph Neural Networks.}
Graph Neural Networks (GNNs) apply iterative message-passing over local neighborhoods and have been effectively used in physical simulation and PDE modeling, notably via the Graph Network-based Simulators (GNS) framework~\cite{sanchez2020learning}. NFTMs generalize this concept from discrete graphs to continuous spatial fields, where read/write heads move and query regions of space rather than fixed graph nodes.

\paragraph{Transformers.}
Transformers~\cite{vaswani2017attention} implement global attention and have been shown to be Turing complete under appropriate conditions~\cite{perez2021finding}. However, their $\mathcal{O}(N^2)$ attention cost and lack of locality bias make them inefficient for large spatial grids. NFTMs instead employ fixed-radius, local neighborhood updates, achieving $\mathcal{O}(N)$ per-step computational cost while retaining full differentiability.

\paragraph{Diffusion Models.}
Diffusion models~\cite{ho2020denoising, song2021scorebased} refine noisy data through iterative denoising, achieving state-of-the-art generative performance. While similar in spirit to NFTMs as iterative solvers, diffusion models rely on stochastic noise injection and score matching, whereas NFTMs employ a deterministic controller that learns local transition rules. This makes NFTMs closer to algorithmic and physical solvers, with diffusion-like refinement emerging as a special case.

\paragraph{Analog Computation Models.}
NFTM also connects to analog models of computation such as the General Purpose Analog Computer (GPAC)~\cite{shannon1941gpac} and the Blum–Shub–Smale (BSS) model~\cite{blum1989real}. Unlike these hand-designed continuous machines, NFTMs are learned from data via gradient descent and integrate seamlessly with modern architectures such as neural fields and differentiable renderers. Thus, NFTM unifies the expressiveness of analog systems with the practicality of trainable deep learning models.

\section{Neural Field Turing Machine: Definition and Universality}
\label{sec:definition}
\begin{figure}
    \centering
    \includegraphics[width=0.99\linewidth]{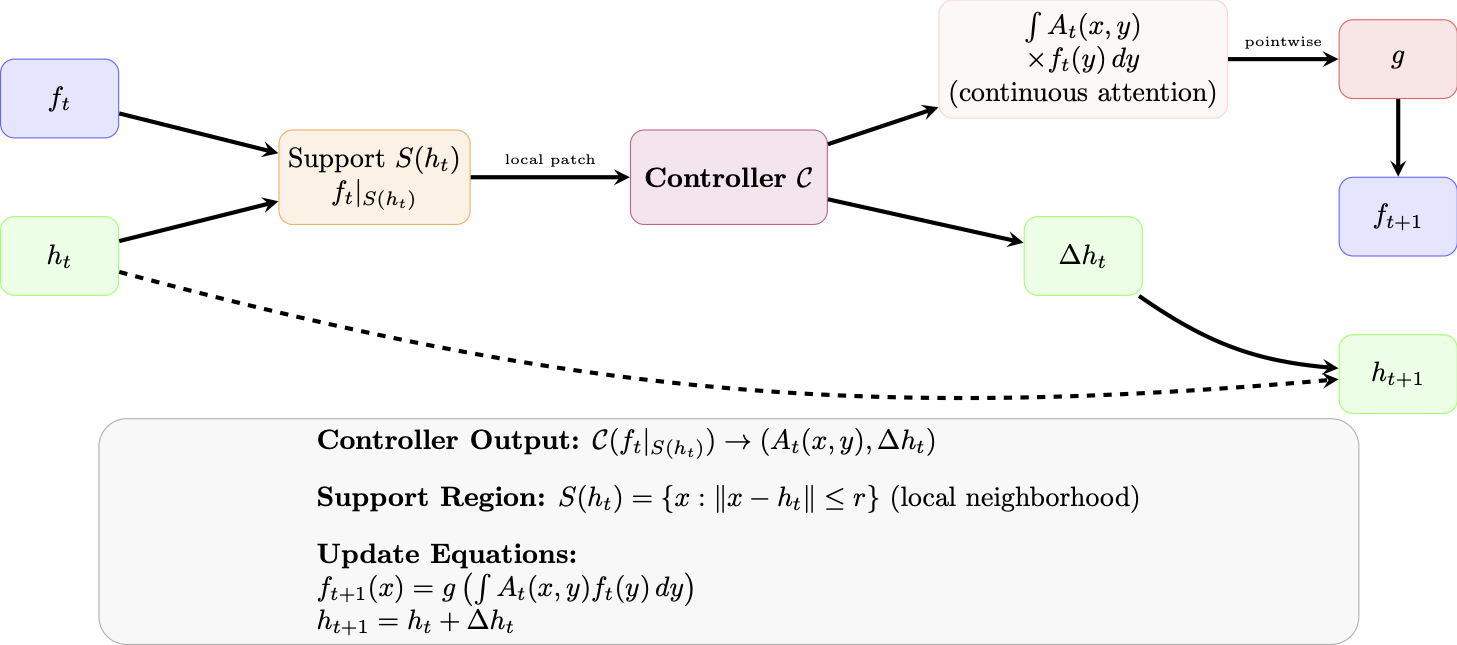}
    \caption{Neural Field Turing Machine (NFTM).}
    \label{fig:nftm}
\end{figure}

A \textit{Neural Field Turing Machine (NFTM)} is a computational model that unifies continuous neural fields with the differentiable read/write mechanisms introduced in Neural Turing Machines. Unlike NTMs, which address a discrete set of memory slots via global attention, NFTMs embed computation directly into a spatial domain, where updates are defined over local neighborhoods and executed by movable read/write heads (see Fig.~\ref{fig:nftm}). This makes NFTM closer in spirit to convolutional or cellular update rules, while retaining the algorithmic flexibility of a controller.

\begin{table}[h]
\centering
\caption{Comparison of NFTM with related models. NFTM combines differentiable read/write from NTMs with the spatial locality of NCAs, while offering an alternative to PINNs for PDE learning.}
\label{tab:comparison}
\resizebox{\linewidth}{!}{%
\begin{tabular}{lcccc}
\toprule
\textbf{Aspect} & \textbf{NTM} & \textbf{NFTM} & \textbf{NCA} & \textbf{PINNs} \\
\midrule
Memory          & Discrete slots & Spatial field & Grid cells & MLP over coords \\
Access          & Global attention & Local heads & Fixed local & PDE residuals \\
Update          & Seq. controller & Euler-style field & Local rule & Residual minimization \\
Domain bias     & Algorithmic & Symbolic, physical, perceptual & Pattern growth & PDE solving \\
Determinism     & Yes & Yes & Yes & Yes \\
Per-step cost   & $\mathcal{O}(N^2)$ & $\mathcal{O}(N)$ & $\mathcal{O}(N)$ & Depends on collocation \\
\bottomrule
\end{tabular}%
}
\end{table}

The system maintains a spatial field $f_t$ at time $t$, which evolves over discrete timesteps. Read/write heads $h_t$ specify positions within this field, and each update is computed over a local support region centered at $h_t$. In the simplest case this is a ball of radius $r$,
\begin{equation}
S(h_t) = \{x \in \mathcal{X} : \|x - h_t\| \leq r\},
\end{equation}
but more generally $S(h_t)$ can be defined by an arbitrary locality kernel $K(x,h_t)$, allowing the region to take non-spherical shapes such as boxes, anisotropic stencils, or ray frustums as in neural rendering. The controller extracts features $f_t[S(h_t)]$ from this region, which serve as the context for computation.

A controller $C$ operates on the local features $f_t[S(h_t)]$ to produce two outputs:
\begin{equation}
C(f_t[S(h_t)]) \mapsto (A_t(x,y), \Delta h_t).
\end{equation}
Here, $A_t(x,y)$ defines a spatial attention field, while $\Delta h_t$ specifies an update to the latent state.  

The update equations governing NFTM dynamics are
\begin{equation}
f_{t+1}(x) = g\left( \int A_t(x,y) f_t(y) \, dy \right),
\end{equation}
\begin{equation}
h_{t+1} = h_t + \Delta h_t,
\end{equation}
where $g$ is a pointwise nonlinearity applied to the aggregated features.  

\begin{proposition}[Turing Completeness of NFTM]
NFTMs are Turing complete under bounded error.
\end{proposition}

The claim follows by reduction to \textit{Rule 110}, a one-dimensional cellular automaton proven to be Turing complete~\cite{cook2004universality}. By discretizing the continuous state $f_t$ into binary values and constraining the support region to a fixed local neighborhood (radius $r=1$), the NFTM controller can replicate the local update rules of Rule 110. Since Rule 110 can emulate any Turing machine, the NFTM inherits this universality. The bounded error condition arises because NFTMs operate in continuous domains, but quantization ensures symbolic dynamics can be recovered with arbitrarily small error.

Thus, the NFTM provides a bridge between Turing-complete symbolic systems and field-based neural dynamics, offering a framework for studying computation in continuous domains with explicit locality and bounded observation.

\section{Methodology}
\label{sec:methodology}

Following the formal definition of NFTM in Section~\ref{sec:definition}, we now show how the model can be instantiated in three domains of increasing complexity: (i) symbolic cellular automata, (ii) physics-informed PDE solvers, and (iii) image inpainting. Across these tasks, the unifying principle is that the controller $\mathcal{C}$ applies local rules at head positions, and emergent global behavior arises through iterative rollout.

\subsection{Cellular Automata Simulation}

We first demonstrate how NFTM can simulate classical cellular automata. 
The controller MLP learns the rule’s truth table and acts on a grid, 
taking as input a cell and its neighborhood, and outputting the cell state at $t+1$: 
\begin{equation}
f_{t+1}(i) = \mathcal{C}\!\left(\text{read\_neighborhood}(f_t, i)\right)
\end{equation}

Here, the number of heads equals the number of cells, 
and the field is discretized so that head positions correspond to cell indices. 
This instantiation is equivalent to Neural Cellular Automata (NCA), 
making NCA a strict subset of NFTM where the field is discretized, heads are fixed, 
and updates are Boolean.  

To maintain differentiability while ensuring discrete states, 
we use Straight-Through Estimator (STE) binarization:
\begin{equation}
\text{STE}(x) = \begin{cases}
\lfloor x + 0.5 \rfloor & \text{(forward)} \\
x & \text{(backward gradient)}
\end{cases}
\end{equation}

With this setup, the controller successfully learns 
Rule 110 and Conway’s Game of Life, both of which are Turing complete.  

\begin{algorithm}[h]
\caption{NFTM Cellular Automata Simulation}
\label{alg:nftm_ca}
\begin{algorithmic}[1]
\State \textbf{Input:} Initial field $f_0$, timesteps $T$, controller $\mathcal{C}$
\State $f \leftarrow f_0$, $\text{fields} \leftarrow [f_0]$
\For{$t = 1$ to $T-1$}
    \State $f_{\text{read}} \leftarrow \text{STE}(f)$
    \State $\text{neighborhoods} \leftarrow \text{read\_neighborhoods}(f_{\text{read}})$
    \State $f \leftarrow \text{STE}(\sigma(\mathcal{C}(\text{neighborhoods})))$
    \State $\text{fields.append}(f)$
\EndFor \\
\Return $\text{fields}$
\end{algorithmic}
\end{algorithm}

\subsection{Physics-Informed Neural PDE Solvers with NFTM}

Beyond symbolic systems, NFTM can act as a neural solver for partial differential equations (PDEs) by combining explicit physics with learnable parameters. This addresses three challenges in physics-informed machine learning: (i) spatial parameter estimation, (ii) stable autoregressive training, and (iii) uncertainty quantification. 

\paragraph{Heteroscedastic Loss.}
To recover spatially varying coefficients $\alpha(x,y)$ we use a heteroscedastic Gaussian likelihood on the local PDE update $\delta$:
\begin{equation}
\mathcal{L}_{\text{NLL}} = \tfrac{1}{2}\frac{(\delta_{\text{gt}} - \alpha \, L_{\text{phys}}(u))^2}{\sigma^2} + \tfrac{\beta}{2}\log \sigma^2,
\end{equation}
where $L_{\text{phys}}$ is the discretized Laplacian and $\sigma^2$ models aleatoric uncertainty. This encourages $\alpha$ to explain systematic variation while $\sigma$ captures residual noise. We optionally reweight by $|L_{\text{phys}}|^\gamma$ to emphasize high-curvature regions, and add weak Gaussian priors on $\alpha$ and $\sigma$ to avoid pathological scales. Positional encodings $[u,x,y]$ break translation equivariance, enabling the recovery of spatially varying $\alpha(x,y)$. 

\paragraph{Two-Phase Training.} 
Autoregressive models face exposure bias: they are trained on ground truth but tested on their own rollouts. We mitigate this via a two-phase scheme. In Phase A (teacher forcing), the controller learns the local operator $\delta = \alpha \nabla^2 u$ using ground truth transitions, which yields stable gradients and rapid parameter identification. In Phase B (rollout training), the NFTM is unrolled from $f_0$ for $T$ steps under its own predictions and optimized with point-wise MSE at head positions plus small regularizers. This enforces stability over long horizons and mitigates exposure bias, forming a curriculum that first learns local physics and then tunes for autoregressive dynamics.

\paragraph{Head-Based Architecture.}
NFTM’s heads collect values and neighbors via a 5-point stencil, apply learned operators, and scatter updates back. This mirrors finite-difference schemes: When the Laplacian is fixed, the the learned $\alpha(x,y)$ retains physical interpretability.

Experiments on 2D heat equations show that heteroscedastic loss recovers $\alpha(x,y)$ with MAE below $0.02$, and two-phase training maintains stability over $50+$ steps with PSNR above $30$ dB.

\subsection{Image Inpainting with NFTM}

NFTM can also tackle perceptual tasks such as image inpainting. Each image is treated as a field, and the controller iteratively refines it by predicting per-pixel corrections given the current image and mask. Training data are drawn from CIFAR-10, normalized to $[-1,1]$, with 25--50\% of pixels randomly removed using a mixture of dropout and square block masks. Corrupted pixels are filled with Gaussian noise, while known pixels are always clamped back to the ground truth.

The controller (a small convolutional stack) outputs a correction $\Delta I_t$, a gating map $g \in [0,1]$, and (optionally) per-channel logvariance for heteroscedastic likelihoods. In this demo we use a homoscedastic Gaussian loss:
\begin{equation}
\mathcal{L}(I, I_{\text{gt}}) = 
\tfrac{1}{2}\Big(\tfrac{(I - I_{\text{gt}})^2}{\sigma^2} + 2\log\sigma\Big) 
+ \lambda_{\text{TV}}\,\mathrm{TV}(I),
\end{equation}
with $\sigma$ fixed and a TV-L1 prior to encourage smoothness. We also include a small contractive penalty $\|I_t - I_{t-1}\|^2$ to discourage oscillations.

To stabilize training, updates are “guarded” by an energy check resembling a line search:
\begin{equation}
I_{t+1} = \mathrm{clamp}\!\left(I_t + \beta \cdot g \cdot \Delta I_t\right),
\end{equation}
where $\mathrm{clamp}(\cdot)$ restores known pixels. If the proposed update increases the energy (data term + TV), the step is rejected and $\beta$ is reduced, ensuring monotonic improvement.

Training follows a curriculum over rollout depth: at each epoch a rollout length $t \in [1,K]$ is sampled, with $K$ gradually increased. Step sizes $\beta$ are annealed during training, and per-step correction magnitudes are clipped with a decay schedule to prevent divergence at long horizons. During evaluation, the model is rolled out for $K_{\text{eval}}$ steps, and performance is measured by PSNR as a function of iteration. 

This setup highlights NFTM’s behavior as an iterative perceptual solver: trained with short rollouts, it generalizes to longer horizons and continues refining images beyond the training depth, unlike fixed-depth feedforward inpainting networks.

\subsection{Summary}

Across these domains, NFTM consistently learns local rules that compose into complex global behaviors. 
In cellular automata, it discovers symbolic truth tables; 
in PDEs, it learns physics-consistent operators; 
in inpainting, it acts as a measurement-consistent solver.  
The design ensures GPU-friendly parallelism (linear in field size and horizon) while bridging discrete algorithmic reasoning and continuous physical simulation within a single framework.

\section{Results}
\label{sec:results}
We evaluate NFTM across three toy demonstrations and one perceptual task, highlighting its ability to unify discrete symbolic rules, continuous PDE dynamics, and iterative perceptual inference within the same framework.

\subsection{Cellular Automata}
\begin{figure}[h]
    \centering
    \includegraphics[width=0.99\linewidth]{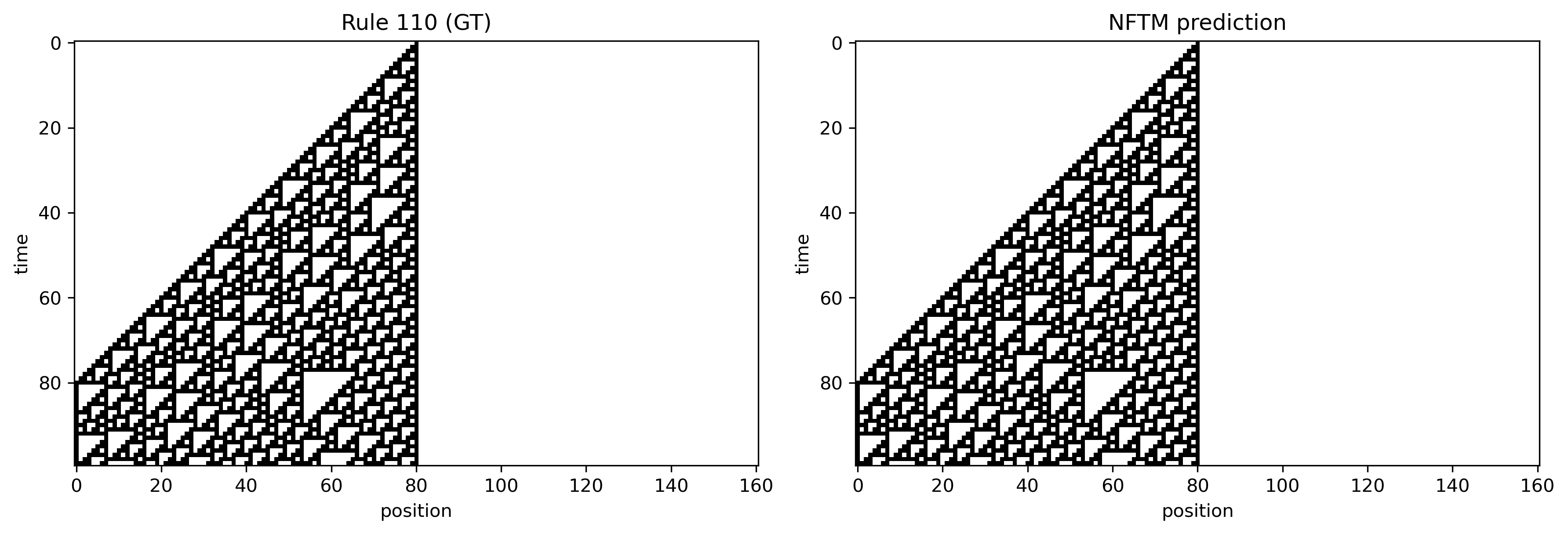}
    \caption{NFTM reproduces Rule~110 cellular automaton dynamics. Left: ground-truth evolution. Right: NFTM prediction. The match over 100 steps demonstrates NFTM’s ability to capture Turing-complete symbolic rules.}
    \label{fig:rule110}
\end{figure}

NFTM with a simple MLP controller successfully learns the local truth table of Rule~110 and reproduces rollouts faithfully. As shown in Figure~\ref{fig:rule110}, NFTM matches the ground-truth cellular automaton evolution over 100 steps when trained on ground-truth rollouts, and crucially continues to extrapolate correctly beyond the horizon of the training sequence. This indicates that the controller has learned the underlying local update rule rather than memorizing trajectories. While the task is ultimately trivial—Rule~110 reduces to a finite truth table—the experiment serves as a minimal validation that NFTM can internalize symbolic update dynamics and reproduce Turing-complete behavior.

\subsection{Heat Equation: Global Diffusion Coefficient}
\begin{figure}[h]
    \centering
    \includegraphics[width=0.99\linewidth]{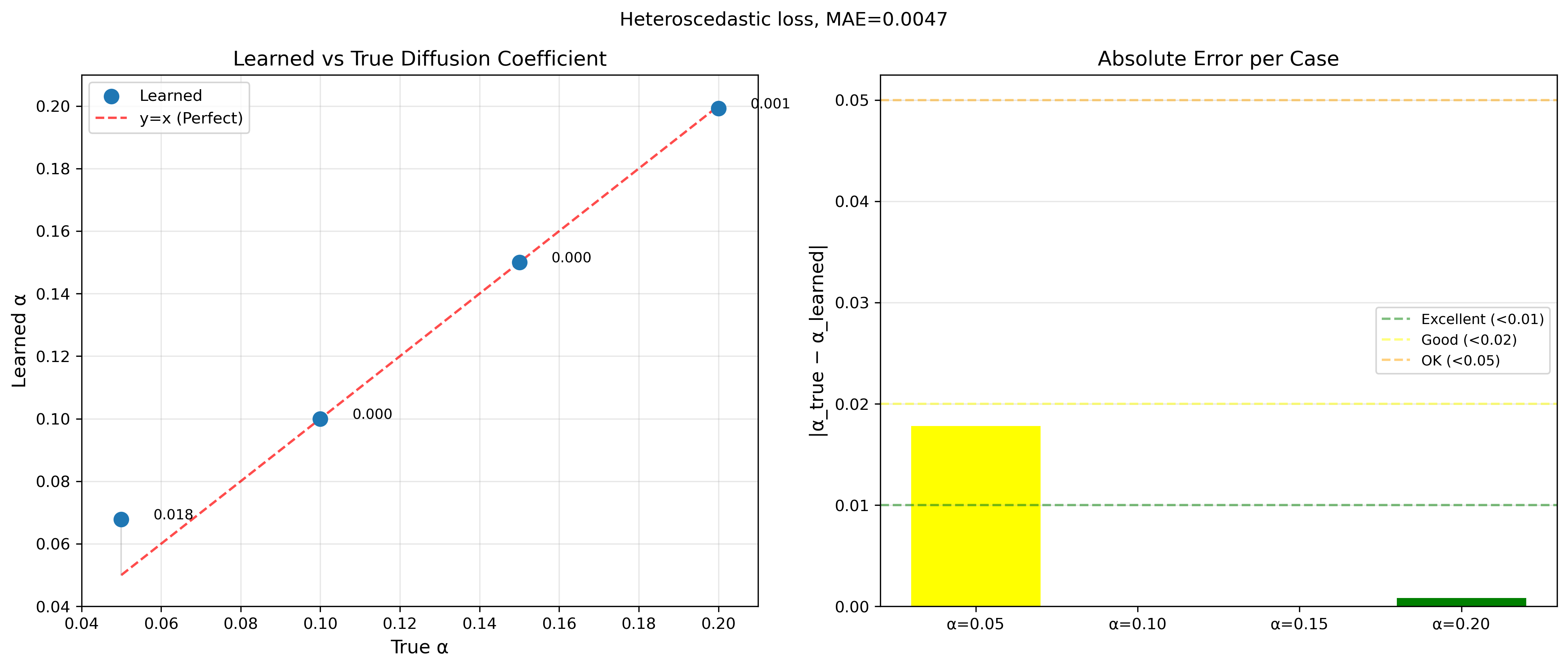}
    \caption{Learning global diffusion coefficient $\alpha$ with heteroscedastic loss. Left: learned vs.~true coefficients. Right: absolute error per case. NFTM recovers $\alpha$ with mean absolute error below $0.01$ in most cases.}
    \label{fig:global-alpha}
\end{figure}

In the global diffusion experiment, NFTM accurately recovers the true diffusion coefficient $\alpha$ using the heteroscedastic loss. As shown in Figure~\ref{fig:global-alpha}, the learned coefficients align closely with the true values across the tested range. For $\alpha = 0.05$, the model learns $0.067$ (absolute error $0.018$). For $\alpha = 0.10$, $0.15$, and $0.20$, the learned values are $0.100$, $0.150$, and $0.200$, respectively, with negligible errors ($\leq 0.001$). This demonstrates that NFTM can reliably infer physical constants from spatio-temporal rollouts.

\subsection{Heat Equation: Variable Diffusion Coefficient}
\begin{figure}[h]
    \centering
    \includegraphics[width=0.99\linewidth]{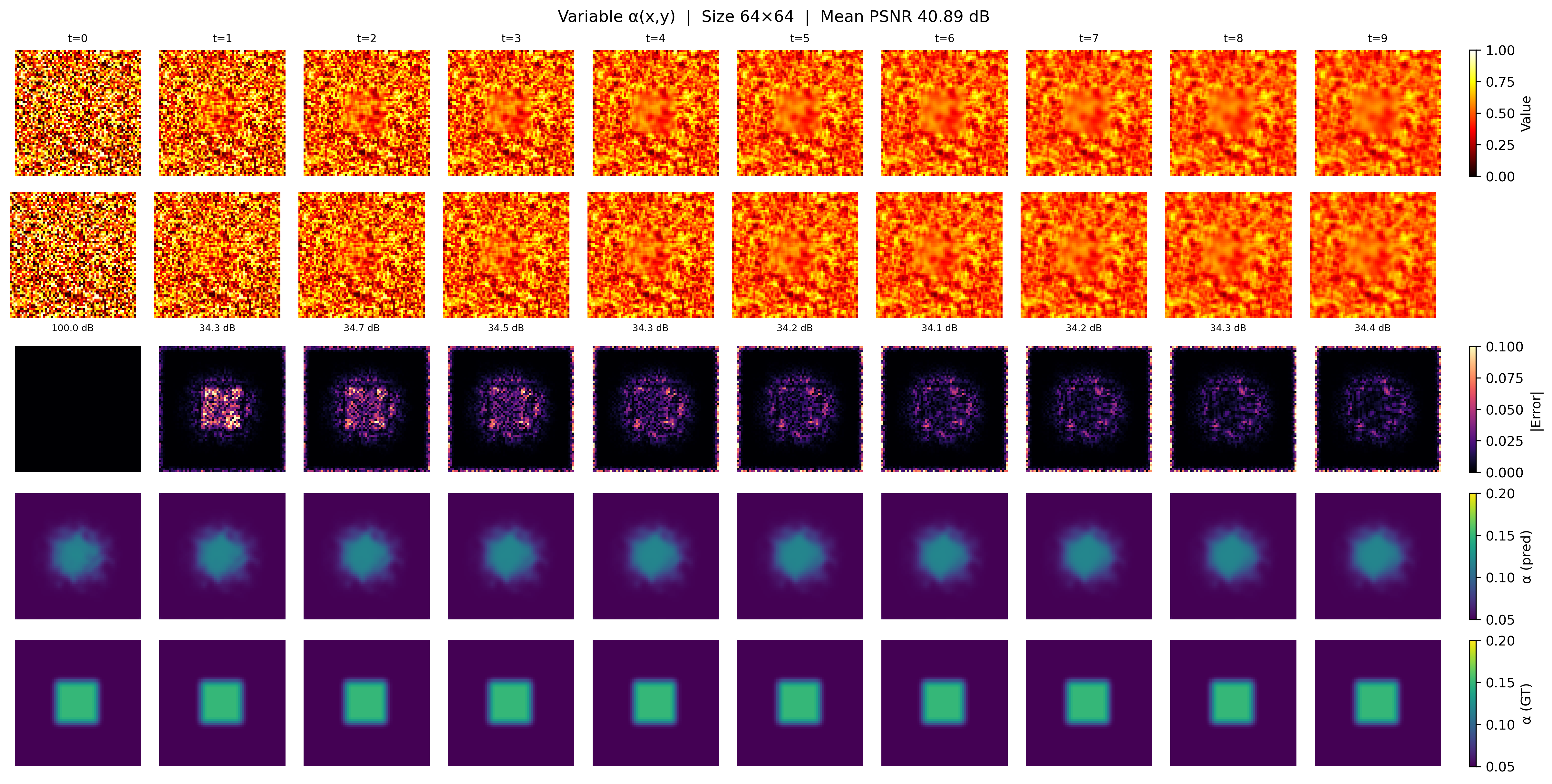}
    \caption{Learning spatially varying diffusion coefficient $\alpha(x,y)$. Top: predicted field over rollout steps. Middle: error maps. Bottom: predicted vs.~ground-truth $\alpha(x,y)$. NFTM accurately reconstructs the spatial variation with mean PSNR $40.89\,\mathrm{dB}$.}
    \label{fig:variable-alpha}
\end{figure}

NFTM also handles spatially varying coefficients $\alpha(x,y)$. In this experiment, $\alpha(x,y)$ was set to $0.05$ everywhere except in a central square region where it was $0.15$, smoothed by three successive $3{\times}3$ average-pool operations. As shown in Figure~\ref{fig:variable-alpha}, NFTM recovers this spatial profile: the predicted $\alpha(x,y)$ closely matches the ground-truth map, with mean PSNR of $40.89\,\mathrm{dB}$. Stepwise evolution shows that prediction error stabilizes around $0.03$–$0.04$ absolute error magnitude after a few steps. These results highlight NFTM’s flexibility: a single architecture can recover both global scalars and spatially varying parameter fields.

\subsection{Image Inpainting}
\begin{figure}[h]
    \centering
    \includegraphics[width=0.7\linewidth]{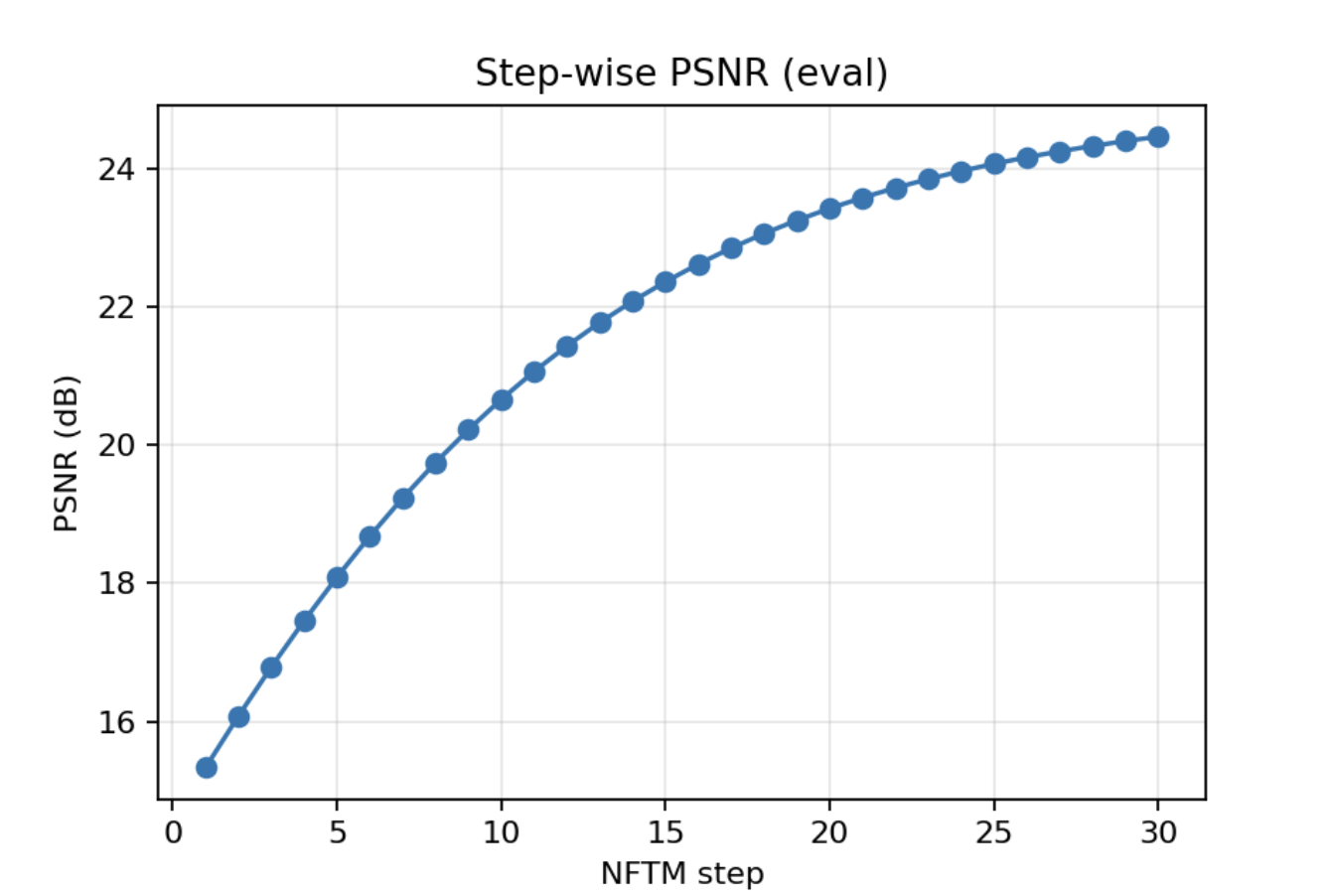}
    \caption{Step-wise average PSNR on CIFAR-10 inpainting task. NFTM exhibits monotonic improvement, generalizing from 20 training steps to 30 evaluation steps.}
    \label{fig:epochs}
\end{figure}

\begin{figure}[h]
    \centering
    \includegraphics[width=0.99\linewidth]{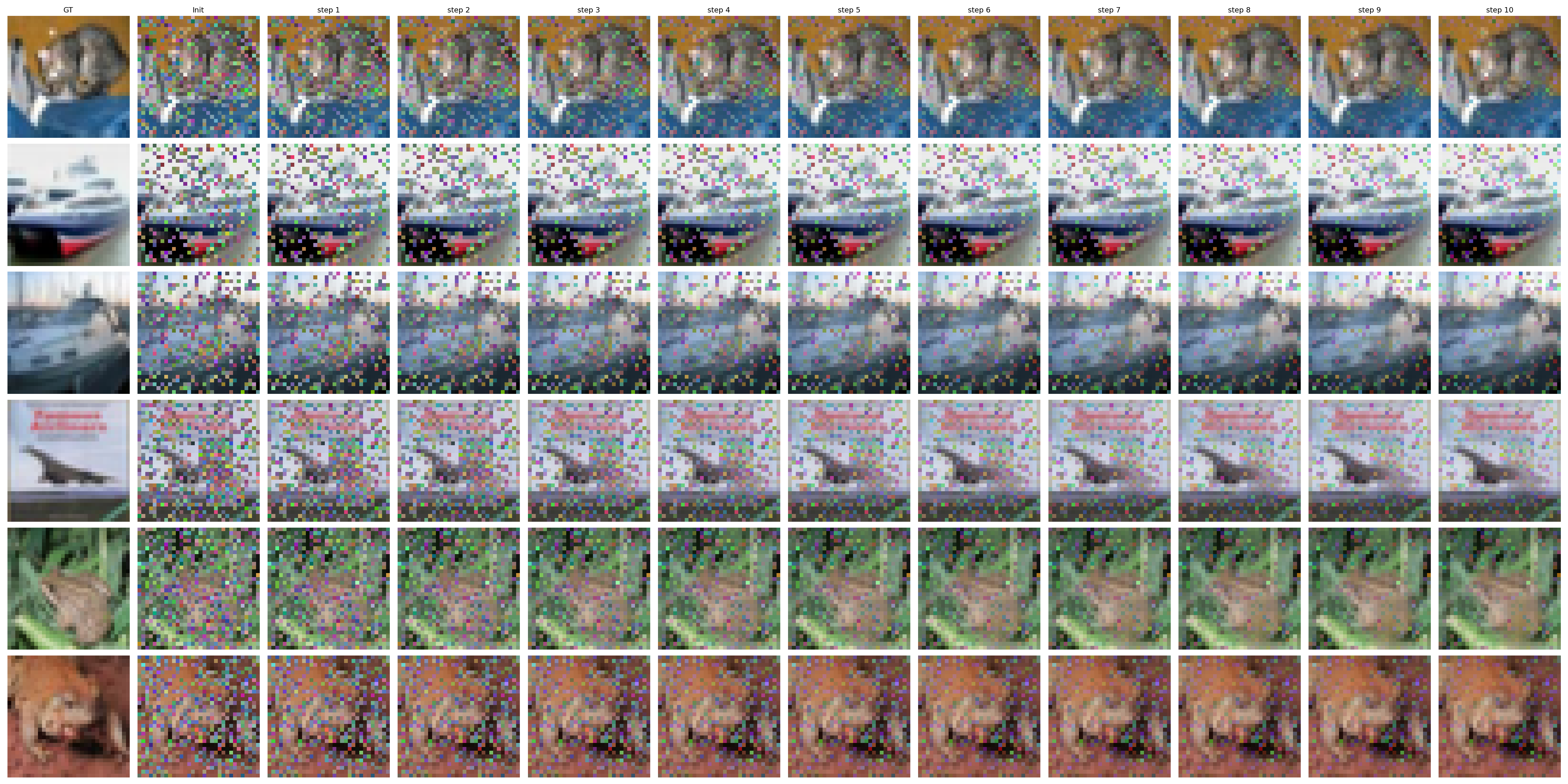}
    \caption{Qualitative image inpainting results on CIFAR-10. Left to right: ground truth, initialization, and successive NFTM refinement steps. NFTM progressively reconstructs occluded regions with higher fidelity.}
    \label{fig:image-interpolation}
\end{figure}

Finally, we apply NFTM to image inpainting on CIFAR-10. The model was trained for 20 rollout steps with guarded backtracking, and evaluated unguarded for 30 steps. As shown in Figure~\ref{fig:epochs}, PSNR improves monotonically throughout inference, rising from $15.2\,\mathrm{dB}$ at step~1 to $24.5\,\mathrm{dB}$ at step~30. Qualitative results (Figure~\ref{fig:image-interpolation}) illustrate how missing regions are progressively reconstructed with increasing fidelity. Importantly, despite training with shorter rollouts, the controller generalizes to longer horizons, continuing to refine the image beyond the training horizon. This extrapolation ability underlines NFTM’s nature as an iterative solver rather than a fixed-depth feedforward model.

\subsection{Discussion}
Across experiments, NFTM demonstrates the ability to learn local update rules that compose into coherent global dynamics. In the heat equation tasks, NFTM adapts seamlessly between learning a single global diffusion coefficient and recovering a spatially varying coefficient field, reflecting its flexibility as a universal computational primitive. In perceptual inpainting, NFTM trained for 20 steps continues to improve quality when rolled out to 30 steps, showing that the controller learns a stable refinement process rather than memorizing a fixed sequence of updates. Together, these findings suggest that NFTM generalizes across problem types (symbolic, physical, perceptual) and across rollout horizons, reinforcing its role as a unifying framework for iterative computation.

Our experiments are designed as illustrative demonstrations of NFTM’s versatility rather than competitive benchmarks. 
For each domain, NFTM subsumes or relates closely to natural baselines: NCAs for cellular automata, CNN/U-Net style solvers for PDEs, and diffusion-based refiners for inpainting. 
While we do not present head-to-head comparisons in this work, our goal is to highlight NFTM’s unifying nature as a differentiable spatial computer capable of spanning symbolic, physical, and perceptual domains. 
We view the present results as proof-of-concept validations, with systematic benchmarking against task-specific architectures left for future work.

\section{Discussion and Conclusion}
\label{sec:conclusion}
A key aspect of NFTM's design is the configurable nature of the support region $S(h_t)$. The framework accommodates variable locality through attention-based selection, with support regions ranging from small local patches ($\mathcal{O}(N)$ complexity) to the entire field ($\mathcal{O}(N^2)$ complexity). NFTM can also employ multiple heads with different support regions simultaneously, enabling multi-scale computation where different heads operate at various spatial resolutions. This allows both fine-grained local processing and global long-range interactions within a single framework, while providing adaptive computation allocation based on spatial complexity. 

NFTM’s local update paradigm also aligns with physical principles: even seemingly long-range forces like gravity and electromagnetism propagate locally through space-time. Iterative local updates can therefore simulate global dynamics by allowing information to propagate gradually. Moreover, the differentiable nature of NFTM opens avenues for enforcing conservation laws. Invertible neural network architectures~\cite{jacobsen2018irevnet,kingma2018glow} can preserve information flow, while geometric deep learning demonstrates how symmetries can be encoded into neural architectures to enforce invariants via Noether’s theorem~\cite{bronstein2021geometric,cohen2016group,finzi2020generalizing}. For example, translation equivariance in the controller could enforce momentum conservation, and rotation equivariance could preserve angular momentum. Embedding such symmetries into NFTM controllers may ensure physically consistent dynamics that respect conservation of energy, momentum, and other invariants.

In terms of computational complexity, NFTM scales linearly with field size $N$ for fixed-radius neighborhoods. For radius $r$, each head processes $\mathcal{O}(r^2)$ elements in 2D (or $\mathcal{O}(r^3)$ in 3D). Since $r$ is constant, the per-head cost is $\mathcal{O}(r^2 + C)$, where $C$ is controller cost, yielding $\mathcal{O}(N)$ overall. This is directly comparable to convolutional networks, which cost $\mathcal{O}(N \times k^2 \times \text{channels})$ for kernel size $k$, or finite-difference solvers that scale with stencil size~\cite{leveque2007finite}. The true trade-off is not asymptotic but constant factors: NFTM’s controller may be more expensive per head than a fixed convolution, but offers far greater flexibility in defining update rules. More significant limitations arise from error accumulation in autoregressive rollouts, a known challenge for sequential prediction models~\cite{bengio2015scheduled}. Our two-phase training reduces exposure bias but does not eliminate it entirely, and long-horizon stability remains sensitive to hyperparameters and discretization. While our experiments span symbolic, physical, and perceptual tasks, they remain at toy scale compared to real-world turbulence, coupled fields, or high-resolution visual reasoning.

Another distinctive property of NFTM is its ability to improve predictions when rolled out beyond its training horizon, effectively trading computation for accuracy. This test-time scaling resembles adaptive computation time~\cite{graves2016adaptive}, where complex inputs are processed with more steps while simple cases converge quickly. Our inpainting experiments confirm that NFTM can act as an iterative solver, refining outputs monotonically over extended rollouts. This property is particularly attractive for spatial reasoning in robotics or perception: a robot could “think longer” in challenging environments while making faster decisions in simple cases. Such adaptive scaling combines the representational power of modern controllers (e.g., U-Nets) with the iterative refinement capabilities of NFTM.

Looking forward, several extensions could enhance NFTM’s capabilities. Extending to 3D fields could impact scientific simulation and perception tasks involving volumetric data or point clouds. Adaptive time-stepping could improve multi-scale dynamics, while physics-informed constraints could enforce conservation directly. Hybrid designs that combine generative models with NFTMs might yield controllable refiners for creative or scientific applications. Most importantly, systematic evaluation on large-scale domains such as climate modeling, materials simulation, or robotics would clarify NFTM’s practical utility relative to established methods.

In summary, we introduced the Neural Field Turing Machine (NFTM), a differentiable architecture that unifies symbolic, physical, and perceptual computation. NFTM combines a spatial field with a neural controller and read/write heads, proving Turing completeness under bounded error and demonstrating versatility across cellular automata, PDE solvers, and image inpainting. Its flexible support regions, capacity for multi-scale reasoning, and ability to trade compute for accuracy suggest NFTM as a step toward a differentiable spatial computer, a unifying framework for iterative field evolution across domains.

\bibliographystyle{plainnat}
\bibliography{references}

\appendix
\appendix
\section{Complexity Analysis of NFTM}
\label{app:complexity}

We analyze the computational complexity of the Neural Field Turing Machine (NFTM) under its three main instantiations: cellular automata simulation, physics-informed PDE solvers, and image inpainting. Across all settings, let $N$ denote the number of spatial sites in the field, $T$ the rollout horizon, and $P$ the computational cost of a forward pass through the controller $\mathcal{C}$.

\subsection{General Complexity}
At each timestep, the NFTM performs three operations:
\begin{enumerate}
    \item Reading a local neighborhood at head positions.
    \item Applying the controller $\mathcal{C}$ to compute updates.
    \item Writing updates back into the field.
\end{enumerate}
For a neighborhood of fixed size $k$, the read/write cost is $\mathcal{O}(kN)$, while the controller adds $\mathcal{O}(NP)$. Thus, the total time complexity is
\[
\mathcal{O}(N \cdot T \cdot (k + P)),
\]
which is effectively linear in field size $N$ and rollout length $T$. Memory usage is $\mathcal{O}(N)$ for forward simulation, and up to $\mathcal{O}(N \cdot T)$ if full trajectories are stored for backpropagation (as in teacher forcing). With truncated backpropagation, this can be reduced to $\mathcal{O}(N \cdot w)$, where $w$ is the truncation window.

\subsection{Local Attention Variant}
If NFTM employs local attention, each head aggregates values from a patch of radius $r$. The cost per timestep becomes
\[
\mathcal{O}(N \cdot r^2 \cdot T),
\]
since each head reads $O(r^2)$ neighbors. For small $r$ (e.g., 1--3), this remains efficient and GPU-parallelizable.

\subsection{Comparison Across Instantiations}
Table~\ref{tab:complexity} summarizes complexity across the three experimental domains.

\begin{table}[h]
\centering
\caption{Asymptotic complexity of NFTM instantiations. $N =$ field size, $T =$ rollout horizon, $P =$ controller cost, $r =$ patch radius.}
\label{tab:complexity}
\begin{tabular}{lcc}
\toprule
\textbf{Task} & \textbf{Complexity} & \textbf{Notes} \\
\midrule
Cellular Automata & $\mathcal{O}(N \cdot T \cdot P)$ & MLP controller, STE binarization \\
PDE Solvers & $\mathcal{O}(N \cdot T \cdot (P+k))$ & 5-point stencil, physics operators \\
Inpainting & $\mathcal{O}(N \cdot T \cdot P)$ & CNN controller, curriculum rollouts \\
NFTM w/ Local Attention & $\mathcal{O}(N \cdot r^2 \cdot T)$ & Patch-based attention, $r$ small \\
\bottomrule
\end{tabular}
\end{table}

\subsection{Discussion}
In all cases, NFTM rollouts scale linearly with field size $N$ and horizon $T$, comparable to classical convolutional neural networks and finite-difference solvers. The constant factors differ by task (e.g., neighborhood size for PDEs, kernel size for CNNs), but remain modest in practice. This ensures that NFTM is both expressive, capable of simulating symbolic and physical systems, and computationally practical for large-scale training on GPUs.

\end{document}